%% file: main.tex
\documentclass[10pt,twocolumn,letterpaper]{article}

\usepackage{iccv}
\usepackage{times}
\usepackage{epsfig}
\usepackage{graphicx}
\usepackage{amsmath}
\usepackage{amssymb}

\usepackage[pagebackref=true,breaklinks=true,letterpaper=true,colorlinks,bookmarks=false]{hyperref}

\iccvfinalcopy 

\def\iccvPaperID{10993} 

\ificcvfinal\fi

\usepackage{import}
\import{misc}{preamble.tex}

\begin{document}

\title{Extended Few-Shot Learning: \\ Exploiting Existing Resources for Novel Tasks}

\author{Reza Esfandiarpoor\\
Brown University\\
{\tt\small re@brown.edu}
\and
Amy Pu\\
Brown University\\
{\tt\small amy\_pu@brown.edu}

\and
Mohsen Hajabdollahi\\
Isfahan University of Tech.\\
{\tt\small m.hajabdollahi@alumni.iut.ac.ir}

\and
Stephen H. Bach\\
Brown University\\
{\tt\small stephen\_bach@brown.edu}
}

\maketitle
\ificcvfinal\thispagestyle{empty}\fi

\begin{abstract}
\import{sections}{abstract.tex}
\end{abstract}

\import{sections}{intro.tex}

\import{sections}{background.tex}
\import{figures}{block_diagram.tex}
\import{sections}{afsl.tex}
\import{sections}{benchmarks.tex}
\import{sections}{learn_aux_data.tex}
\import{sections}{selection.tex}

\import{sections}{embedding.tex}
\import{figures}{visual_images.tex}
\import{sections}{masking.tex}
\import{tables}{imagenet_main.tex}

\import{sections}{classification.tex}
\import{sections}{experiments.tex}
\import{tables}{cifar_main.tex}

\import{sections}{results.tex}

\import{tables}{benchmark_table.tex}
\import{tables}{transfer.tex}

\import{sections}{analysis.tex}

\import{figures}{mask_improvement.tex}

\import{sections}{related_works.tex}

\import{sections}{conclusion.tex}

\section*{Acknowledgements}

This material is based on research sponsored by Defense Advanced Research Projects Agency (DARPA) and Air Force Research Laboratory (AFRL) under agreement number FA8750-19-2-1006, and by the National Science Foundation (NSF) under award IIS-1813444. The U.S. Government is authorized to reproduce and distribute reprints for Governmental purposes notwithstanding any copyright notation thereon. The views and conclusions contained herein are those of the authors and should not be interpreted as necessarily representing the official policies or endorsements, either expressed or implied, of Defense Advanced Research Projects Agency (DARPA) and Air Force Research Laboratory (AFRL) or the U.S. Government. We gratefully acknowledge support from Google and Cisco. Disclosure: Stephen Bach is an advisor to Snorkel AI, a company that provides software and services for weakly supervised machine learning.

{\small
\bibliographystyle{ieee_fullname}
\bibliography{misc/iccv_bib}
}

\clearpage
\import{appendix}{appendix.tex}

\end{document}

%% file: misc/preamble.tex
\usepackage{multirow}
\usepackage{caption}
\usepackage{subcaption}
\usepackage[sort, numbers]{natbib}
\usepackage{placeins}
\usepackage{lipsum}
\usepackage{layouts}
\usepackage[dvipsnames]{xcolor}
\usepackage{soul}
\usepackage[capitalise]{cleveref}
\crefname{section}{\S}{\S\S}
\Crefname{Section}{\S}{\S\S}
\usepackage{booktabs}
\usepackage{comment}

\newcommand{\cifar}{CIFAR-100}
\newcommand{\cifarfs}{CIFAR-FS}
\newcommand{\fc}{FC-100}
\newcommand{\mini}{\textit{mini}ImageNet}
\newcommand{\tiered}{\textit{tiered}ImageNet}

%% file: sections/abstract.tex
In many practical few-shot learning problems, even though labeled examples are scarce, there are abundant auxiliary datasets that potentially contain useful information.
We propose the problem of extended few-shot learning to study these scenarios.
We then introduce a framework to address the challenges of efficiently selecting and effectively using auxiliary data in few-shot image classification.
Given a large auxiliary dataset and a notion of semantic similarity among classes, we automatically select pseudo shots, which are labeled examples from other classes related to the target task.
We show that naive approaches, such as (1)
modeling these additional examples the same as the target task examples or (2) using them to learn features via transfer learning, only increase accuracy by a modest amount.
Instead, we propose a masking module that adjusts the features of auxiliary data to be more similar to those of the target classes.
We show that this masking module performs better than naively modeling the support examples and transfer learning by 4.68 and 6.03 percentage points, respectively. Code is available at: \href{https://github.com/BatsResearch/efsl}{https://github.com/BatsResearch/efsl}.

%% file: sections/intro.tex
\section{Introduction}
\label{sec:intro}

Large labeled datasets for novel machine learning tasks are costly and time-consuming to gather, so there is a great demand for methods that can learn new tasks with limited labeling.
One such area of work is few-shot learning (FSL).
In FSL, the model must predict the labels of query images given only one or a few labeled examples of each target class, called the support set.
Recent FSL works limit the available training resources to the support set and a labeled set of base classes~\cite{vinyals2016matching, snell2017prototypical, santoro2016meta, mishra2017simple, finn2017model, lee2019meta}.
While the traditional FSL setting is an important test of generalization from limited resources, we argue that it fails to capture many other realistic scenarios found in practice.
Although the number of base classes is small (hundreds), the  size and breadth of labeled datasets publicly accessible and internally available to organizations are great.
In scenarios where parts of those massive datasets are semantically related to novel tasks, the challenge is not limited labeled data per se, but a lack of labeled examples of the novel classes.

\import{figures}{ven_diagram.tex}

Now, practitioners face new questions.
Given a novel few-shot task, \emph{which} labeled data should be used as additional information?
\emph{How} should these related examples be incorporated into the model?
For example, consider an online advertising platform that needs to identify images of a recently introduced shoe.
In the typical FSL setup, the company has to train a model with just a few available images of that specific shoe.
The typical approach ignores the company’s database of previous models, as well as their semantic knowledge about the properties of this new shoe (\eg, available colors, style, target customers, etc.).
There are numerous other similar situations in industry, government, and research, where proprietary datasets and knowledge bases, public ones, or combinations of both are available when trying to solve new FSL tasks. 
From another perspective, humans use a significant amount of prior knowledge and data to solve novel tasks.
Therefore, it is important to thoroughly investigate how machine learning models can use these resources in low-label regimes like FSL.

We introduce \textbf{E}xtended \textbf{F}ew-\textbf{S}hot \textbf{L}earning (EFSL) as a formulation of this problem.
In EFSL, models can take advantage of other available resources outside of labeled examples of the target classes, in addition to the support set.
As illustrated in~\Cref{fig:ven_diagram}, EFSL harnesses a wider range of available resources compared to similar problems.

We contribute benchmarks for this new problem, which are designed to evaluate EFSL models with auxiliary data of different semantic distances to the target task.
We organize a large auxiliary dataset such as ImageNet22k~\cite{imagenet_dataset} in a hierarchy  and consider different scenarios in which the target classes, some number of ancestors, and all their descendants are removed from the auxiliary data.
These benchmarks provide a fair evaluation of future work on EFSL by controlling the contribution of auxiliary data.

The EFSL task introduces several new challenges, which we address with a novel framework.
First, it is essential to intelligently select a subset of auxiliary data that is related to the target task.
We propose using common sense knowledge graphs (e.g., ConceptNet~\cite{speer2016conceptnet}) for selecting the most related auxiliary classes to target classes.
We call this related subset of auxiliary examples \emph{pseudo shots}.
Our method takes advantage of the direct relation between semantic and visual similarity of images~\citep{deselaers2011visual}.
Second, it is necessary to design an adaptable mechanism to effectively incorporate pseudo shots into existing FSL models. Pseudo shots often contain noisy and irrelevant information, because they are examples of classes distinct from the target classes. We find that naively using FSL methods on pseudo shots fails to fully exploit the potential of auxiliary data, especially as the pseudo shots' semantic distance increases from the target classes.
To address this issue, we design a masking module that identifies the more helpful information.
We train the masking module in a meta-learning fashion through numerous EFSL episodes to identify the most helpful similarities between auxiliary data and the target task.
We also show that this framework is compatible with many traditional FSL classifiers including nearest centroid classifiers~\cite{chen2020new, snell2017prototypical, xing2019adaptive, gidaris2018dynamic, oreshkin2018tadam, triantafillou2017few}, relation networks~\cite{sung2018learning}, and embedding adaptation methods~\cite{ye:cvpr20}, demonstrating the versatility of auxiliary data in the FSL setting.

Finally, we conduct extensive experiments to study various aspects of the EFSL problem.
We find that the magnitude of improvement for image classification crucially depends on the modeling choices made, especially when the auxiliary data is semantically distant from the target classes. When auxiliary data is used only to train a better image embedding function, average accuracy increases by 5.86 percentage points.

Further using auxiliary data to naively update target class prototypes reduces the average gain to 4.13 percentage points.
In contrast, using our proposed masking module increases average accuracy by a total of 8.81 percentage points relative to the baseline without any auxiliary data.

The improvement provided by the masking module over basic feature embeddings increases from 4.68 to 12.85 points as the semantic difference between the auxiliary data and target classes grows. Our findings indicate that employing auxiliary data is a challenging problem requiring specific models designed to address the questions specific to this new problem.

We summarize our main contributions:
\begin{itemize}
    
    \item
    We propose the EFSL problem to study methods for learning to exploit existing resources in practical low-shot regimes.
    
    \item
    We propose benchmarks for EFSL that systematically control the semantic distance between target classes and available auxiliary data.
    
    \item
    We introduce a method for auxiliary data selection that takes advantage of semantic information to select the most informative subset of auxiliary data.
    
    \item
    We design a masking module that learns to identify similarities between auxiliary data and target classes, in order to filter irrelevant information.
    
    \item
    We conduct extensive experiments to analyze the challenges of the EFSL problem. 
    We find that straightforward applications of existing techniques are suboptimal, and that methods designed for EFSL can significantly improve accuracy.

\end{itemize}

%% file: figures/ven_diagram.tex
\begin{figure}[t]
\centering
\includegraphics[width=0.9\columnwidth]{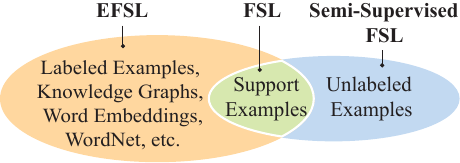}

\caption{The extended few-shot learning (EFSL) setting in relation to traditional few-shot learning (FSL) and semi-supervised FSL.
All settings use support examples of the target classes. Semi-supervised FSL adds unlabeled examples.
In contrast, EFSL uses external labeled data, disjoint from support classes, to harness the link between non-visual sources and labels.}

\label{fig:ven_diagram}
\end{figure}

%% file: sections/background.tex
\section{Background}
\label{sec:background}
In traditional FSL, the model is trained and evaluated on a set of base classes $C^{train}$ and testing classes $C^{test}$, respectively.
Note that base classes and testing classes are disjoint, \ie, $C^{train} \cap C^{test}=\emptyset$.
Traditional FSL follows an episodic scheme~\citep{vinyals2016matching, snell2017prototypical, santoro2016meta, mishra2017simple, finn2017model, lee2019meta}.
To create a K-shot N-way episode, we randomly select a set of N classes as support, \ie target, classes $C$.
Based on the \textbf{few-shot assumption}, there are only $K$ labeled images available per support class, where $K$ is an small number.
Thus, we randomly select $K$ images from each of the $N$ classes.
This set of $N \times K$ examples is known as the support set, $S$.
We take another $q$ samples from each of the $N$ classes as our query set, $Q$.
The model is aware of the support set labels and is expected to predict the query set labels.
Testing and training episodes are created in a similar way, but, the support classes are selected from $C^{test}$ and $C^{train}$, respectively.

%% file: figures/block_diagram.tex
\begin{figure*}[t]
\begin{center}
\includegraphics{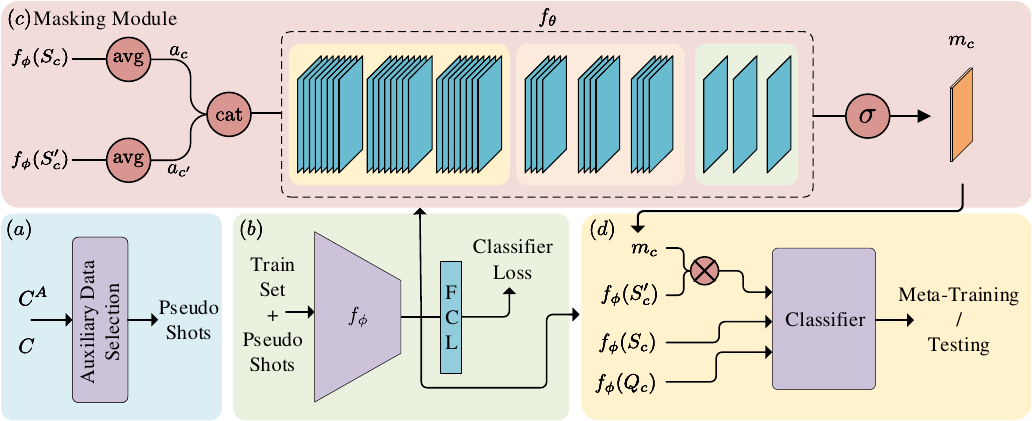}
\end{center}
   \caption{
   \textbf{The proposed framework for EFSL.}
   a) We use common sense knowledge graphs to select a subset of auxiliary examples.
   b) We use the base classes and the corresponding pseudo shots to train the feature embedding module.
   c) The masking module takes the support set and pseudo shot embeddings and generates a mask $m_c$ to filter the irrelevant information in pseudo shots.
   d) We feed the query, support set, and masked pseudo shot embeddings to a classifier and generate the logits for meta-training/testing.
   }
\label{fig:block_diagram}
\end{figure*}

%% file: sections/afsl.tex
\section{Extended Few-Shot Learning (EFSL)}
\label{sec:afsl_definition}
Extended few-shot learning has a similar goal to traditional FSL, \ie, predicting the query labels given a small support set.
However, in EFSL, the input is extended beyond the support set and includes other labeled datasets and \emph{non-visual}, semantic knowledge sources.
Examples of such datasets are public benchmark data and proprietary datasets belonging to organizations.
External labeled datasets are constrained to exclude all examples of the support classes, but there is no constraint on the non-visual resources.
Importantly, EFSL preserves the few-shot assumption in traditional FSL, defined in~\cref{sec:background}, because there are no examples of support classes in auxiliary resources.

Given the immense size and variety of available auxiliary datasets, one of  EFSL's goals is to find a subset of useful and informative examples.
EFSL extends the traditional episodes by an auxiliary support set $S'_c$ which consists of the $K'$ most informative auxiliary examples for each class $c$.
Formally, the extended episodes are as following
\begin{equation} \label{eq:augmented_episode}
    \mathcal{E}'_c = \mathcal{E}_c \cup \{S'_c\} = \{ S_c, S'_c, Q_c \} \quad \forall c \in C \; .
\end{equation}

%% file: sections/benchmarks.tex
\subsection{EFSL Benchmarks}
\label{sec:benchmarks}
Here, we introduce benchmarks to evaluate EFSL models.
We intend to evaluate scenarios with varying degrees of similarity between the auxiliary data and target task.
Ideally, an algorithm would do well even with distantly related data, and, if that is not possible, degrade gracefully.
To create auxiliary datasets of increasing semantic distance from the target tasks, we use the hierarchical relationships among words in WordNet~\citep{wordnet} to prune data.
In WordNet, organizing words with respect to the hyponym (subtype) relation creates a directed tree in which child classes are subtypes of their parents.
In all scenarios, we eliminate class $c$ and all of its descendants, and use the remaining classes as the set of allowed auxiliary classes with respect to $c$.
As we select data of increasing semantic distance, we move up the tree and prune ancestors of the class $c$ and their descendants.
We refer to each such dataset as a level from $0$ (pruning class $c$ and its descendants) to $3$ (pruning the great-grandparents of class $c$ and all its descendants).

Formally, let $p^c_l$ be the $l^{th}$ ancestor of class $c$ such that $p^c_0 = c$ and $p^c_n$ is the root of the WordNet tree where $n$ is the depth of class $c$ in the tree.
Also, let $d^c$ be the set of all the descendants of class $c$.
Given a set of classes, $\mathbb{C}$, we constrain the auxiliary classes at level $l$ and use $C^A$ as the set of available auxiliary classes with respect to $\mathbb{C}$
\begin{equation} \label{eq:benchmarks}
    C^A = C^T \setminus \bigcup_{c \in \mathbb{C}} d^{p^c_l} \; ,
\end{equation}
where $C^T$ is the set of all the classes in the auxiliary data source.
In our experiments, we use this technique to extend several FSL benchmarks.

%% file: sections/learn_aux_data.tex
\section{Learning with Auxiliary Data}

Here, we address the main challenges of EFSL. \Cref{fig:block_diagram} shows the block diagram of our framework.
In~\cref{sec:selection}, we tackle auxiliary data selection using a common sense knowledge graph~(\Cref{fig:block_diagram}a). 
Next, taking in the support and auxiliary sets, we first try a simple approach for designing a feature embedding module in~\cref{sec:embedding}~(\Cref{fig:block_diagram}b).
However, the performance of these basic embeddings is mediocre, with performance often being harmed by semantically distant auxiliary data (\cref{sec:experiments} and \cref{sec:analysis}).
We address this deficiency in~\cref{sec:masking} with our masking module, which filters the irrelevant information in auxiliary data~(\Cref{fig:block_diagram}c).
The resulting masked embeddings are compatible with a wide range of existing FSL classification techniques, demonstrating the extensive applicability of exploiting auxiliary data.
In~\cref{sec:classifier}, we use three popular methods for standard FSL to classify the embeddings: nearest centroid classifier~\citep{tian2020rethinking, chen2020new, snell2017prototypical, sung2018learning}, relation networks~\citep{sung2018learning}, and embedding adaptation~\citep{ye:cvpr20}~(\Cref{fig:block_diagram}d).
We demonstrate that all three methods significantly benefit from the higher quality masked embeddings. 

%% file: sections/selection.tex
\subsection{Auxiliary Data Selection}
\label{sec:selection}
Our selection method intends to maximize the semantic and visual similarities between the support classes and selected auxiliary examples.
We can view these distantly related auxiliary examples as low quality instances of the support classes. 
Thus, we call such examples \emph{pseudo shots}.
In~\cref{sec:analysis}, we show that not only does our selection of auxiliary data contain more useful information than random examples, but also that random examples usually lead to poor performance.

We propose using a notion of semantic similarity between two classes to select a subset of auxiliary examples.
Semantic similarity has a broader applicability and is computationally more efficient than visual similarity.
Visual similarity requires pair-wise image comparisons with high computational cost.
Because semantic and visual similarity between two categories are directly related~\citep{deselaers2011visual},
we can avoid expensive pair-wise image comparisons by using semantic similarity as a computationally efficient proxy for visual similarity.
This setup can potentially discover a wider visual range of auxiliary data that pertains to each support class.

To estimate the semantic similarity between two classes, we employ \emph{common sense knowledge graphs}, graph structures modeling natural concepts and their relations.
We focus on the ConceptNet graph~\citep{speer2016conceptnet}, given its wide domain coverage of concepts, multiple relationship types, and useful Euclidean word embeddings to capture semantic similarity.
These embeddings combine the relational knowledge in ConceptNet with distributional semantics from Word2Vec~\citep{mikolov:arxiv13} and GloVe~\citep{pennington:emnlp14} embeddings using a generalized retrofitting method~\citep{faruqui2014retrofitting}.
This hybrid semantic space is highly useful in word relatedness evaluations, with nodes having a smaller distance not just to those with names that often co-occur in text, but also those that are related in other ways, such as having similar functions, properties, appearing in similar contexts or physical locations, etc.

We use a metric of cosine similarity between two classes' corresponding node vectors in ConceptNet.
For class $c$, we select the $N'$ most similar auxiliary classes $C'_c$, and randomly choose $K'$ samples from them as the auxiliary support set $S'_c$:
\begin{equation}
C'_c = \operatorname{argmax}_{\{c_1, \dots, c_{N'}\} \subseteq C^A} \sum_{c_i} \langle E_c, E_{c_i} \rangle \; , 
\end{equation}
where $C^A$ is the set of available auxiliary classes, $E$ is the corresponding word vector, and $\langle \cdot \rangle$ is cosine similarity.

%% file: sections/embedding.tex
\subsection{Feature Embedding}
\label{sec:embedding}
We use an embedding function $f_\phi$ to represent images as lower-dimensional feature vectors.
We train a ResNet-12 on the union of base classes and pseudo shots, selected by following~\cref{sec:selection}.
Following previous works~\citep{tian2020rethinking, chen2020new}, we  use a typical cross entropy loss.
After training, we drop the fully connected output layer and use the remaining model as our embedding function $f_\phi$. We freeze the parameters of $f_\phi$ after this.
We refer to the features generated by $f_\phi$ as basic embeddings.

%% file: figures/visual_images.tex
\begin{figure}[t]
\centering
\includegraphics[width=0.9\columnwidth]{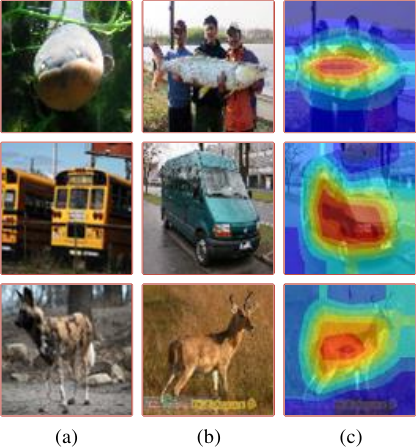}

\caption{Support images, pseudo shots, and masked pseudo shots for three different classes are shown in columns (a), (b), and (c), respectively. Warmer colors represent higher mask values. The masking module successfully learns to identify the most similar region to support images in pseudo shots.}
\label{fig:visuals}
\end{figure}

%% file: sections/masking.tex
\subsection{Masking Module}
\label{sec:masking}
Although basic embeddings improve the accuracy, as shown by our experiments in~\cref{sec:results}, their performance is mediocre, especially when the pseudo shots and auxiliary classes are semantically distant.
To maximize our gain, we design a masking module that compares the features of the support set and pseudo shots, and learns to filter the irrelevant information in pseudo shots. 
 
To avoid computationally expensive pairwise comparisons between support set and pseudo shots, we compare the prototype of each support class with the prototype of the corresponding pseudo shots.
For each support class $c$, we use a mean aggregator over basic embeddings to calculate the prototypes.
Let basic embeddings, the output of $f_\phi$, be of shape $(C^*, W, H)$ where $C^*$, $W$, and $H$ are the number of channels, width, and height of the embeddings, respectively.\footnote{{To avoid confusion with the set of support classes, we use $C^*$ rather than the conventional $C$ to denote the number of channels.}}
In each episode, the mean aggregator transforms the dimensions as 
\begin{equation} \label{eq:mean}
\operatorname{mean}: (N, K^*, C^*, W, H) \rightarrow (N, C^*, W, H) \; ,
\end{equation}
where $K^*$ equals $K$ and $K'$ for support images and pseudo shots, respectively.

We define a function $f_\theta$ that takes two sets of feature embeddings as input, and outputs a 2D matrix with similar dimensions to feature maps.
Then, we use the sigmoid function to calculate matrix $m$ with values in [0, 1].
Ideally, $m$ assigns 1 to the regions where both input embeddings contain identical information and 0 to regions with totally dissimilar information and a number between 0 and 1 for other regions with varying degree of similarity.

We use $f_\theta$ to identify the similarities between support set and pseudo shot prototypes.
For each class $c$, we concatenate the support set prototype $a_c$ and pseudo shot prototype $a'_c$ calculated by~\cref{eq:mean}. Let $m_c$ be the mask for class $c$
\begin{gather*}
m_c = \sigma(f_\theta(\operatorname{cat}(a_c, a_{c'}))) \\
\sigma(f_\theta(.)): (N, 2C^*, W,H) \rightarrow (N, 1, W, H) \; .
\end{gather*}

To exploit the inherent spatial correlation of images, we calculate a spatial mask for the $(H, W)$ plane rather than having a single scalar mask value per $(H, W)$ plane.
We update pseudo shot embeddings by element-wise multiplication of $m_c$ with each of the $C^*$ feature maps.
We refer to these updated features maps as masked embeddings.

\Cref{fig:visuals} shows images of support classes, their corresponding pseudo shot, and the generated mask. The masking module successfully identifies the similarities between images and filters less useful information.

%% file: tables/imagenet_main.tex
\begin{table*}[t]
  \centering
  \begin{tabular}{lccccccc}
    \toprule
    
    Model && \multicolumn{2}{c}{\mini{}} && \multicolumn{2}{c}{\tiered{}} 
   \\
    \midrule
    && 1-shot & 5-shot && 1-shot & 5-shot \\

\cmidrule(r){3-4} \cmidrule(r){6-7}

NCC Baseline   && $ 56.7 \pm 0.71 $ & $ 75.93 \pm 0.53 $ && $ 64.46 \pm 0.83 $ & $ 82.21 \pm 0.62 $ \\
NCC + Basic PS  && $ 69.93 \pm 0.76 $ & $ 76.70 \pm 0.64 $ && $ 76.09 \pm 0.81 $ & $ 82.38 \pm 0.67 $ \\
NCC + Masked PS  && $\mathbf{ 75.82 \pm 0.62 }$ & $\mathbf{ 85.93 \pm 0.42 }$ && $\mathbf{ 80.66 \pm 0.68 }$ & $\mathbf{ 87.83 \pm 0.51 }$ \\

\midrule
RM Baseline && $  58.42 \pm  0.77  $  &  $  74.42  \pm  0.55  $  &&  $  62.58  \pm  0.85  $  &  $  78.65  \pm  0.69  $ \\
RM + Basic PS && $ 66.67 \pm 0.78 $ & $ 72.39 \pm 0.65 $ && $ 64.39 \pm 0.88 $ & $ 72.62 \pm 0.81 $ \\
RM + Masked PS && $\mathbf{ 67.48 \pm 0.72 }$ & $\mathbf{ 78.67 \pm 0.53 }$ && $\mathbf{ 67.51 \pm 0.85 }$ & $\mathbf{ 78.70 \pm 0.69 }$ \\

\midrule
EA Baseline   && $ 61.69 \pm 0.74 $ & $ 77.24 \pm 0.52 $ && $ 67.92 \pm 0.84 $ & $ 82.90 \pm 0.59 $ \\
EA + Basic PS && $ 72.42 \pm 0.72 $ & $ 80.58 \pm 0.58 $ && $ 75.52 \pm 0.80 $ & $ 81.94 \pm 0.69 $ \\
EA + Masked PS  && $\mathbf{ 76.88 \pm 0.67 }$ & $\mathbf{ 86.28 \pm 0.41 }$ && $\mathbf{ 80.54 \pm 0.69 }$ & $\mathbf{ 87.53 \pm 0.53 }$ \\

    \bottomrule
  \end{tabular}
  \captionsetup{justification=centering}
\caption{EFSL results on \mini{} and \tiered{} with level 0 pruning of the auxiliary data. Results are for three classifier types: nearest-centroid classifier (NCC), relation module (RM), and embedding adaptation (EA). Each section shows results for the baseline without any auxiliary data, the baseline with basic embeddings of auxiliary data, and the proposed method using masked embeddings. Each entry is mean accuracy over 800 episodes with 95\% confidence intervals.}
\label{tab:ilsvrc_main}
\end{table*}

%% file: sections/classification.tex
\subsection{Classifier}
\label{sec:classifier}
After filtering extraneous information, we can view masked pseudo shots as weaker examples of the support classes.
Thus, we merge the pseudo shots and support set features
\begin{equation}\label{eq:final_supportset}
    S_c^f = \{f_\phi(x); \; \forall x \in S_c    \} \cup \{f_\phi(x) \odot m_c; \; \forall x \in S'_c    \} \; ,
\end{equation}
where $\odot$ is the Hadamard product and $S_c^f$ is the merged support set for class $c$.
The structure of the merged episode is similar to a regular FSL task, allowing us to use a wide range of regular FSL classifiers for prediction.
We evaluate the merged task with three popular FSL classifiers: nearest centroid classifier, relation module, and embedding adaptation.
See the appendix for more details on classifiers and architecture of the masking module.

\textbf{Nearest centroid classifier (NCC).}
NCC is the most popular classifier in FSL literature~\cite{chen2020new, snell2017prototypical, xing2019adaptive, gidaris2018dynamic, oreshkin2018tadam, triantafillou2017few}.
For each class $c$, the centroid $h_c$ is the average over the merged support set $S_c^f$.
For a query example  $x$,~\cref{eq:predicted_dist} calculates a probability distribution over support classes
\begin{equation}\label{eq:predicted_dist}
p(y = c | x) = \frac{\operatorname{exp}(\langle f_\phi(x), h_c \rangle)}{\sum_{c' \in C}\operatorname{exp}(\langle f_\phi(x), h_{c'} \rangle)} \quad \forall c \in C\; ,
\end{equation}
where $\langle \cdot \rangle$ is cosine similarity.
The query example is labeled with the highest probability class.

\textbf{Relation Module (RM).}
Relation networks consist of a feature embedding module and a relation module for classification~\citep{sung2018learning}.
The relation module generates a relation score for each pair of class centroid and query example features. For each query example, we calculate the relation score for all classes and select the class with highest relation score.

\textbf{Embedding adaptation (EA).} 
FSL models usually use the same embedding function for all tasks.
Several works have proposed adapting the feature embeddings for each FSL task to be more discriminative~\citep{hou2019cross, li:cvpr19, ye:cvpr20}.
Similar to~Ye et al.~\citep{ye:cvpr20}, we calculate the centroids for all support classes, then use a Transformer function~\citep{vaswani2017attention} to modify the centroids for the current episode. With the updated class centroids, we use~\cref{eq:predicted_dist} to calculate the probability distribution for support classes and choose the most likely class.

\textbf{Training.} We train our pipeline incrementally.
First, we train the feature embedding function and freeze its parameters.
We then train the masking module with the nearest centroid classifier through numerous episodes.
Next, we replace NCC with another classifier, freeze the parameters of both feature embedding and masking module and repeat the same episodic training. 

%% file: sections/experiments.tex
\section{Experimental Evaluation}
\label{sec:experiments}
In this section, we evaluate the proposed framework for the EFSL problem. 
As expected, adding auxiliary resources to the input improves the performance.
More importantly, we find that addressing challenges like auxiliary data selection and designing specific models for EFSL have a significant impact on performance.
Our masking module improves the accuracy by 4.68 points compared to naively using basic auxiliary embeddings.
Later, in~\cref{sec:analysis}, we find that not addressing these challenges can even cause failure resulting in  accuracy lower than standard FSL.

\subsection{Datasets}
\label{subsec:dataset}
We extend four popular FSL datasets and create benchmarks for EFSL using the procedure from~\cref{sec:benchmarks}.
\textbf{\mini{}} and \textbf{\tiered{}} are both subsets of the ILSVRC-12 dataset~\citep{ilsvrc}.
\mini{}~\citep{vinyals2016matching} consists of 100 classes with 600 images per class.
Classes are split into 64, 20, and 16 classes for training, testing, and validation, respectively.
\tiered{}~\citep{ren2018meta} contains 608 classes, of which 351 classes are for training, 160 for testing, and 97 for validation.
Both \mini{} and \tiered{} contain $84 \times 84$ RGB images.
\textbf{\cifarfs{}} and \textbf{\fc{}} are both variants of the \cifar{} dataset for FSL.
\cifarfs{}~\citep{lee2019meta} splits \cifar{} into 64, 20, and 16 classes, and 
\fc{}~\citep{oreshkin2018tadam} splits \cifar{} into 60, 20, and 20 classes for training, testing, and validation, respectively.
Both datasets contain $32 \times 32$ RGB images.

For auxiliary data, we choose classes with more than 500 samples as the set of auxiliary classes $C^T$ from  \textbf{ImageNet22k}~\citep{imagenet_dataset}.
For each test episode, we use support classes $C$ for $\mathbb{C}$  in~\cref{eq:benchmarks} to get the allowed classes $C^A$ for level $l$ auxiliary data.
For training, we use all of the test classes $C^{test}$ for $\mathbb{C}$ in~\cref{eq:benchmarks} to prune $C^T$ and enforce the desired semantic distance with all test classes.

%% file: tables/cifar_main.tex
\begin{table*}[t]
  \centering
  \begin{tabular}{lccccccc}
    \toprule
    
    Model && \multicolumn{2}{c}{\cifarfs{}} && \multicolumn{2}{c}{\fc{}} 
   \\
    \midrule
    && 1-shot & 5-shot && 1-shot & 5-shot \\

\cmidrule(r){3-4} \cmidrule(r){6-7}

NCC Baseline  && $ 66.56 \pm 0.71 $ & $ 84.43 \pm 0.56 $ && $ 40.28 \pm 0.59 $ & $ 56.25 \pm 0.59 $ \\
NCC + Basic PS  && $ 75.78 \pm 0.79 $ & $ 85.79 \pm 0.54 $ && $ 49.51 \pm 0.67 $ & $ 57.51 \pm 0.66 $ \\
NCC + Masked PS  && $\mathbf{ 82.00 \pm 0.65 }$ & $\mathbf{ 90.49 \pm 0.49 }$ && $\mathbf{ 53.72 \pm 0.67 }$ & $\mathbf{ 65.52 \pm  0.62 }$ \\

\midrule
RM Baseline && $  67.81  \pm  0.80  $  &  $  82.26  \pm  0.64  $  &&  $  37.51  \pm  0.62  $  &  $  50.28  \pm  0.60  $ \\
RM + Basic PS&&  $ 73.32 \pm 0.84 $ & $ 81.68 \pm 0.64 $ && $\mathbf{ 43.22 \pm 0.65 }$ & $ 49.18 \pm 0.66 $ \\
RM + Masked PS &&  $\mathbf{ 74.17 \pm 0.80 }$ & $\mathbf{ 86.21 \pm 0.57  }$ && $ 42.04 \pm 0.65  $ & $\mathbf{ 52.02 \pm 0.63  }$ \\

\midrule
EA Baseline   && $ 69.86 \pm 0.74 $ & $ 85.32 \pm 0.57 $ && $ 41.52 \pm 0.62 $ & $ 57.46 \pm 0.64 $ \\
EA + Basic PS  && $ 77.74 \pm 0.77 $ & $ 87.08 \pm 0.53 $ && $ 51.23 \pm 0.65 $ & $ 58.13 \pm 0.65 $ \\
EA + Masked PS  && $\mathbf{ 83.02 \pm 0.62 }$ & $\mathbf{ 90.84 \pm 0.48 }$ && $ \mathbf{ 54.59 \pm 0.66 }$ & $\mathbf{ 65.64 \pm 0.63 }$ \\

    \bottomrule
  \end{tabular}
  \captionsetup{justification=centering}
\caption{EFSL results on \cifarfs{} and \fc{} with level 0 pruning of the auxiliary data. Results are for three classifier types: nearest-centroid classifier (NCC), relation module (RM), and embedding adaptation (EA). Each section shows results for the baseline without any auxiliary data, the baseline with basic embeddings of auxiliary data, and the proposed method using masked embeddings. Each entry is mean accuracy over 800 episodes with 95\% confidence intervals.}
\label{tab:cifar_main}
\end{table*}

%% file: sections/results.tex
\subsection{Results}
\label{sec:results}
We evaluate using both basic and masked embeddings to create centroids, and consider all three classification methods from~\cref{sec:classifier} on the four benchmarks with level 0 auxiliary data.
We emphasize the results at level 0, which is still a challenging setting because the auxiliary data contains no samples of the test classes nor any subtypes of the test classes.
For context, we also report the accuracy of the classifiers for traditional FSL.
These baselines help better understand the benefits of different approaches to EFSL and underscore the different effects of auxiliary data at different levels of semantic distance.

\textbf{ImageNet.}
\Cref{tab:ilsvrc_main} reports the accuracy for the three classifiers on ImageNet derivatives.
Adding pseudo shots with basic embeddings improves the accuracy for all combinations of the models, tasks, and datasets, with a greater impact for 1-shot tasks.
This consistent improvement shows the value of auxiliary resources in low-shot regimes. 
Using masked pseudo shot embeddings further improves the performance of basic embeddings across all experiments.
On average, the masking module boosts the accuracy of the basic embeddings by 3.98 and 6.39 points for 1- and 5-shot tasks, respectively with the maximum improvement of 9.23 points.
Compared to traditional FSL, these existing resources improve the average accuracy by a total of 12.85 and 5.60 points for 1- and 5-shot tasks, respectively.

\textbf{\cifar{}.}
In~\Cref{tab:cifar_main}, we present the accuracy of the classifiers on variants of \cifar{}.
As expected, adding auxiliary data with basic embeddings improves the performance for almost all tasks.
Consistent with previous results, using masked embeddings further improves the accuracy of basic embeddings.
On average, the masking module improves the accuracy of the basic embeddings by 3.12 and 5.23 points for 1- and 5-shot tasks, respectively.
The masking module increases the accuracy of basic embeddings by up to 8.01 points.
The masking module boosts the average accuracy of the traditional FSL by 11 and 5.79 points for 1- and 5-shot tasks, respectively.

%% file: tables/benchmark_table.tex
\begin{table}[b]
  \centering
  \begin{tabular}{lccccc}
    \toprule
    
    Model & Level & \multicolumn{2}{c}{\mini{}} & \multicolumn{2}{c}{\tiered{}} 
   \\
    \midrule
    && 1-shot & 5-shot & 1-shot & 5-shot \\

\cmidrule(r){3-4} \cmidrule(r){5-6}

Baseline   && $ 61.69 $ & $ 77.24 $ & $ 67.92 $ & $ 82.90 $ \\

\midrule

\multirow{4}{*}{Basic} & 0 &
$ 72.42 $ & $ 	80.58 $ & $ 	75.52 $ & $ 	81.94 $ \\

&1&
$ 70.58 $ & $ 	76.98 $ & $ 	71.74 $ & $ 	80.51 $ \\

&2&
$ 66.67 $ & $ 	75.74 $ & $ 	65.99 $ & $ 	76.22 $ \\

&3&
$ 60.02 $ & $ 	71.81 $ & $ 	63.67 $ & $ 	75.84 $ \\

\midrule

\multirow{4}{*}{Masked} & 0 &		
$ 76.88 $ & $ 	86.28 $ & $ 	80.54 $ & $ 	87.53 $ \\

&1&
$ 75.43 $ & $ 	83.98 $ & $ 	80.54 $ & $ 	87.53 $ \\

&2&
$ 73.79 $ & $ 	83.84 $ & $ 	73.29 $ & $ 	86.36 $ \\

&3&
$ 68.73 $ & $ 	80.36 $ & $ 	73.27 $ & $ 	86.24 $ \\

    \bottomrule
  \end{tabular}
\caption{Embedding adaptation classifier results on \mini{} and \tiered{} with level 0--3 auxiliary data. Results are for the baseline without any auxiliary data, the baseline with basic embeddings of auxiliary data, and the proposed method using masked embeddings. Each entry is mean accuracy over 800 episodes.}
\label{tab:benchmarks}
\end{table}

%% file: tables/transfer.tex
\begin{table}[t]
  \centering
  \begin{tabular}{lcccc}
    \toprule
    
    Model & \multicolumn{2}{c}{\mini{}} & \multicolumn{2}{c}{\tiered{}} 
   \\
    \midrule
    & 1-shot & 5-shot & 1-shot & 5-shot \\

\cmidrule(r){2-3} \cmidrule(r){4-5}

Traditional FSL &
$ 56.7  $ & $ 	75.93  $ & $ 	64.46  $ & $ 	82.21 $ \\

\midrule

Pre-training &
$ 59.81  $ & $ 	82.38  $ & $ 	72.07  $ & $ 	87.19 $ \\

Fine Tuning &
$ 62.71 $ & $ 	77.29 $ & $ 72.51 $ & $ 82.64 $ \\

\midrule

NCC + Masked PS &
$75.82$ & $85.93$ & $80.66$ & $87.83 $ \\

    \bottomrule
  \end{tabular}
\caption{EFSL results on \mini{} and \tiered{} with level 0 auxiliary data. Results are for the nearest centroid classifier, with various ablations and alternative setups. Accuracies are over 800 episodes.}
\label{tab:transfer}
\end{table}

%% file: sections/analysis.tex
\section{Analysis}
\label{sec:analysis}
Here, we further study the auxiliary data with basic and masked embeddings and the impact of semantic distance on performance. We compare EFSL with transfer learning and semi-supervised FSL, and discuss their differences.

\textbf{Auxiliary Data Benchmarks.}
\Cref{tab:benchmarks} reports the performance of the embedding adaptation classifier with different levels of auxiliary data.
We clearly see the value of auxiliary data as masked embeddings improve over the traditional FSL accuracy by 4.71 points even with level 3 auxiliary data--which contains no WordNet great-grandparents of the test classes nor their subtypes.

The accuracy of the masked embeddings degrades more gracefully than basic embeddings with more semantically distant auxiliary examples, by 5.66 and 9.78 points, respectively.
As shown in~\Cref{fig:mask_improvement}, the improvement of masked embeddings over basic embeddings increases for higher levels of auxiliary data.
The masking module thus provides a solution to the real world problem when auxiliary data similar to the target classes are not easily obtainable.

\textbf{Proposed Framework vs. Transfer Learning.}
Transfer learning aims to reuse pre-trained representations for new tasks. Models can use pre-trained parameters as-is or fine tune on new tasks. We use our pre-trained embedding function to evaluate the pre-training performance. We also fine-tune our pre-trained embedding function with test pseudo shots to compare an alternative approach. As shown in~\Cref{tab:transfer}, pre-training improves the traditional FSL accuracy. Fine-tuning on test pseudo shots improves the 1-shot task but harms the 5-shot compared to only pre-training. Our masking method surpasses pre-training and fine-tuning accuracy by 6.03 and 7.15 points, respectively. Evaluating 800 episodes with our masking method is much faster than fine-tuning, 5 minutes vs. 15 hours.
We also show that our method gains benefit from using auxiliary data at test time even without pre-training benefits. See appendix for details.

\textbf{EFSL vs. Semi-Supervised FSL.}
Semi-supervised FSL (SSFSL) is the closest problem to EFSL, but still has significant differences.
Compared to traditional FSL, SSFSL models can use unlabeled examples both during training and testing.
Labeled data encodes valuable knowledge from human annotators, but SSFSL misses this source of knowledge.
Without labels, it is not possible to link visual data and semantic knowledge sources.
Restricted to unlabeled data, SSFSL cannot access the large sources of rich knowledge (\eg, knowledge graphs, knowledge bases, word embeddings, to name a few).
In general, EFSL can use the auxiliary examples more effectively with the extra metadata that is available.
We try to use SSFSL methods for solving EFSL, but we find that the performance of SSFSL methods is sub-optimal and we need models designed specifically for EFSL. Experimental results are provided in the appendix.

\textbf{Random Auxiliary Data.} We compare the performance of our selection method,~\cref{sec:selection}, with random sampling. We find that random sampling fails catastrophically with basic embeddings. Using masked embeddings improves the accuracy but it is still far behind our proposed selection method. Results are reported in the appendix.

%% file: figures/mask_improvement.tex
\begin{figure}[t]
\centering

\begin{subfigure}{0.45\columnwidth}
\centering
\includegraphics[width=\textwidth]{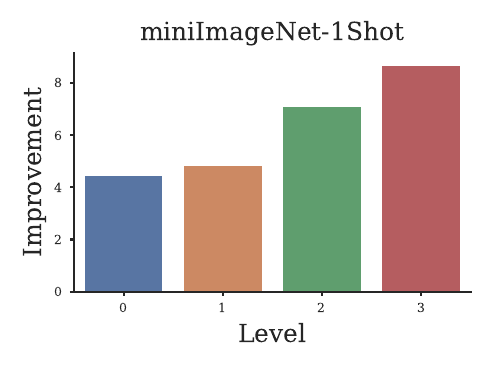} 
\end{subfigure}
\begin{subfigure}{0.45\columnwidth}
\centering
\includegraphics[width=\textwidth]{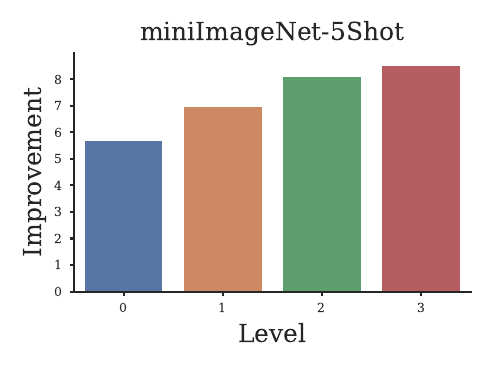} 
\end{subfigure}

\begin{subfigure}{0.49\columnwidth}
\centering
\includegraphics[width=\textwidth]{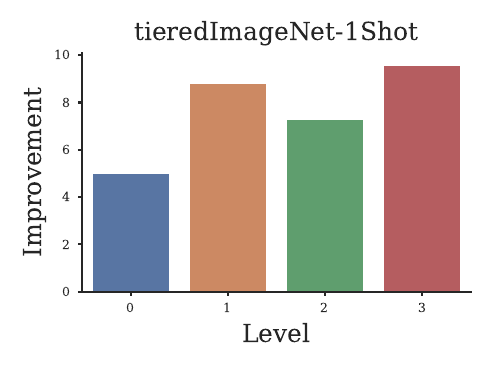} 
\end{subfigure}
\begin{subfigure}{0.49\columnwidth}
\centering
\includegraphics[width=\textwidth]{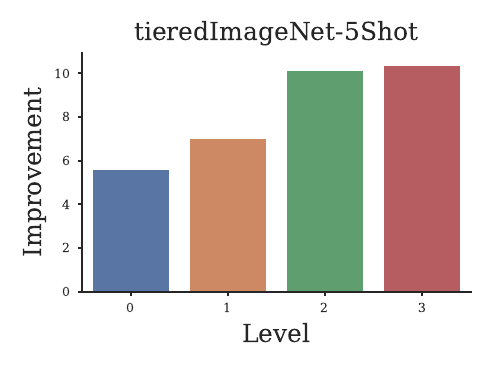} 
\end{subfigure}

\caption{The improvement of masked over basic embeddings for level 0 to 3 auxiliary data, ordered left to right.
}
\label{fig:mask_improvement}
\end{figure}

%% file: sections/related_works.tex
\section{Related Work}
\label{sec:related_works}
Meta-learning is the dominant approach for solving FSL problems. Meta-learning tries to learn transferable knowledge based on the training classes and use this learned knowledge during test time. This transferable knowledge can be a discriminating metric space~\citep{vinyals2016matching, snell2017prototypical, sung2018learning, oreshkin2018tadam, koch2015siamese}, or a fast converging algorithm or initial state~\citep{finn2017model, rusu2018meta, lee2019meta, ravi2016optimization, nichol2018first}. All of these methods rely solely on the support set.

Transductive models have recently gained attention for FSL. These models use the query set as unlabeled data in each episode~\citep{liu2018learning, kim2019edge}.
EGNN~\citep{kim2019edge} uses a graph convolutional edge labeling network to propagate the support set labels to query samples. TPN~\citep{liu2018learning} propagates the labels of the support set to unlabeled (query) samples by learning to construct a graph structure. There are a number of other transductive FSL solutions that rely on the query set as well as the support set in each episode~\citep{hu2020leveraging, garcia2017few}.

Semi-supervised FSL methods are similar to transductive methods, but their unlabeled set is not the same as the query set~\citep{yu:cvpr20, ren2018meta, gidaris:iccv19, simon:cvpr20, lin:cvpr20}. Ren \etal~\citep{ren2018meta} include unlabeled samples in each episode and propose a masking mechanism to control the effect of unrelated unlabeled samples. As the source of unlabeled samples, they use a mix of samples from random and support classes with a 1:1 ratio.
Several other works~\cite{simon:cvpr20, lin:cvpr20, liu2018learning} use the same unlabeled set as Ren \etal~\cite{ren2018meta}.
TransMatch~\cite{yu:cvpr20} draws unlabeled samples from support classes in each episode and measures robustness against distractors, but does not try to exploit them.
Gidaris \etal~\cite{gidaris:iccv19} use \tiered{} as the source of unlabeled data for self-supervised learning for \mini{}.

Some recent studies aim for more discriminative features and propose related masking methods that further refine the embedded features~\citep{hou2019cross, li:cvpr19}. CTM~\cite{li:cvpr19} looks at all the support samples together and generates a mask that indicates the most discriminative features for the current task. CAN~\citep{hou2019cross} calculates a cross attention map for each pair of class centroid features and query features. The cross attention map improves the discriminative power of features by localizing the target object.
Our masking module differs from these works by comparing support sets and pseudo shots to import information from the pseudo shots.

Other related work includes Xing \etal~\cite{xing2019adaptive}, which uses word vectors directly to update class prototypes, whereas we use word vectors as a tool to select auxiliary samples.
Ge and Yu~\cite{ge:cvpr17} use visual similarity and Zhang \etal~\citep{zhang:eccv18} use meta-learning to select auxiliary data for fine-grained image classification.
Finally, concurrent work~\cite{bateni2020improving} considers transductive FSL for multi-domain problems.

%% file: sections/conclusion.tex
\section{Conclusion}
\label{sec:conclusion}
In this work, we introduced the EFSL problem and associated benchmarks to capture the many practical scenarios in which auxiliary resources can aid in novel few-shot tasks.
We proposed a framework for EFSL that uses semantic knowledge to aid the selection of auxiliary data and a masking module to select the useful parts of that data.
It is compatible with a wide range of existing methods for FSL.
We showed that it outperforms naive solutions and is robust as the available auxiliary data grows semantically distant from the test classes.
We believe that the problem of exploiting auxiliary data for new tasks will be increasingly important as shared datasets continue to multiply.

%% file: appendix/appendix.tex
\appendix

\section{Further Analysis}

\textbf{Proposed Framework without Pre-training.}
To study the benefits of using pseudo shots at test time, we train the embedding function using only base classes and repeat our main experiments.
As shown in~\cref{tab:transfer_part2}, using pseudo shots at test time improves the performance of the traditional FSL even without pre-training.
This indicates that the combination of our selection method and masking module adds useful information at test time, supporting our claim that benefits of pseudo shots are beyond simple pre-training.

\import{tables}{transfer_part2.tex}

\textbf{Semi-Supervised FSL methods for EFSL.}
Semi-supervised FSL methods use unlabeled examples as auxiliary data. We study the performance of SSFSL methods on EFSL tasks. We select the masking soft K-means method of Ren \etal~\citep{ren2018meta} (\textbf{MS K-Means)}.
We choose this specific semi-supervised method because it has a distinct masking module that we can integrate in our framework.
It is a soft masking mechanism for updating class centroids.
This mechanism masks entire images based on the assumption that the unlabeled data is a mix of target classes and distractors.
In contrast, our proposed masking module masks individual feature patches of images based on the assumption that the auxiliary data contains related images, but no actual samples of the target class.
To have a fair comparison, we implement MS K-Means with our (higher capacity) embedding function trained on the same data.
We freeze the embedding function and apply their masking mechanism on the resulting embeddings.
With this implementation, MS K-Means is identical to our pipeline other than their masking module.

The performance of MS K-Means with level 0 auxiliary data is reported in~\Cref{tab:ssfsl_random}.
We select the auxiliary data following~\cref{sec:selection}.
The proposed masking module performs significantly better than MS K-Means for both 1-shot tasks, where the limited label problem is severe.
Our masking method also performs better for 5-shot tasks, especially on the smaller \mini{}.
It emphasizes the need for designing models to specifically address EFSL challenges.
\import{tables}{all_analysis.tex}

\textbf{Random Auxiliary Data.}
We ablate the proposed selection method in~\cref{sec:selection} and use random sampling to measure its impact in our framework.
As shown in~\Cref{tab:ssfsl_random}, random sampling fails catastrophically with basic embeddings. As expected, the masking module removes a significant amount of the irrelevant information and increases the accuracy. But, even with masking, random sampling performs poorly on 1-shot tasks and is far behind the proposed selection method on 5-shot tasks.
\import{tables}{imagenet_all_levels.tex}

\textbf{Auxiliary Data Benchmarks.}
We evaluate the performance of all three classifiers with level 0--3 auxiliary data.
\Cref{tab:imagenet_all_levels} reports the mean accuracy on \mini{} and \tiered{}.
The mean accuracy on \cifarfs{} and \fc{} is presented in~\Cref{tab:cifar_all_levels}.
As the semantic distance between target classes and pseudo shots increases, the accuracy decreases.
Models with masked embeddings degrade more gracefully than models with basic embeddings.
The masking module extracts useful information even from level 3 auxiliary data.
It consistently improves the performance of traditional FSL for \mini{} and \tiered{} with level 3 auxiliary data.
For \cifarfs{} and \fc{}, the performance of masked pseudo shots is comparable to traditional FSL.
The masking module fails to extract useful information for \cifar{} derivatives from small $32 \times 32$ images at level 3 auxiliary data.
It is likely that the limited spatial information in smaller images reduces the benefits of a spatial mask.

\section{Architecture}

\textbf{Embedding Function.}
In all cases, we use ResNet12~\citep{he2016deep} as our embedding function with DropBlock~\citep{ghiasi:neurips18} for regularization.
We use (640, 320, 160, 64) filters instead of the original (512, 256, 128, 64) filters in ResNet12.
With this modification, our embedding function is similar to that of Tian \etal~\cite{tian2020rethinking}.

\textbf{Masking Module.}
We achieve the spatial mask introduced in~\cref{sec:masking} with several ResBlocks~\citep{he2016deep} arranged in a pyramid structure for $f_\theta$. Specifically, it consists of three ResBlocks with $(2C^*, C^*, 1)$ filters. In our implementation, the three ResBlocks have (640, 320, 1) filters.

\section{Optimizer}
We use stochastic gradient descent (SGD) as optimizer with initial learning rate of 0.05, momentum of 0.9, weight decay of 5.e-4, and learning rate decay of 0.1.
We train the embedding function for 100 epochs and decay the learning rate at epochs 60 and 80.
We use the parameters at the last epoch as $f_\phi$.
We train the masking module and classifiers for 150 epochs and decay the learning rate at epoch 70.
We use the parameters at the epoch with highest validation accuracy for the masking module and the classifiers.

\import{tables}{cifar_all_levels.tex}

\section{Dataset Alignment.}
ImageNet classes are WordNet IDs.
Thus, it is straightforward to map classes to WordNet nodes.
To map the classes to ConceptNet nodes, we use the corresponding noun in WordNet database for each class.
\cifar{} classes are nouns.
We use all the corresponding WordNet IDs to the selected support classes as $\mathbb{C}$ in~\cref{eq:benchmarks}.
We directly use nouns for mapping \cifar{} classes to ConceptNet nodes.

\section{Classifiers}

\textbf{Relation Module.}
As explained in~\cref{sec:classifier}, relation network consists of two major modules: a feature embedding module and a relation module~\citep{sung2018learning}. The relation module $f_\beta$ is parameterized with convolutional layers followed by several linear layers. The relation module takes the concatenated embeddings of the query example and class centroid and outputs a single scalar. Using a sigmoid non-linearity, the relation module outputs a relation score in [0, 1]. The query example is labeled with the class with highest relation score. Let $h_c$ be the centroid of class $c$ calculated as 
\begin{equation}
h_c = \frac{\sum_{x \in S_c^f}  x}{|S_c^f|} \quad \forall c \in C \; .
\end{equation}
For query example $x$
\begin{equation}
    r_c = \sigma(f_\beta(\operatorname{cat}(h_c, f_\phi(x))) \quad \forall c \in C \; ,
\end{equation}
where $r_c$ is the relation score.

\textbf{Embedding Adaptation.}
Most FSL methods train one embedding function with base classes.
After training the embedding function, all tasks use the same parameters, \ie, the embedding function is task agnostic.
Embedding adaptation models customize the feature embeddings for each task such that they are more discriminative~\citep{hou2019cross, li:cvpr19, ye:cvpr20}. Here, we use the FEAT (Few-shot Embedding Adaptation with Transformer) model from~\citep{ye:cvpr20}. Let $T$ be a set-to-set function that takes the set of support class centroids as input.
Then, $T$ returns an adapted set of cluster centroids, that are more discriminative for the current task. Formally, $H^T = T(H)$ where $H$ is the set of support class centroids, $H = \{h_c: \;  c \in C\}$.
For a query example $x$, we use $H^T$ in~\cref{eq:predicted_dist} to calculate the probability distribution over support classes.

%% file: tables/transfer_part2.tex
\begin{table}[b]
  \centering
  \begin{tabular}{lcccc}
    \toprule
    
    Model & \multicolumn{2}{c}{\mini{}} & \multicolumn{2}{c}{\tiered{}} 
   \\
    \midrule
    & 1-shot & 5-shot & 1-shot & 5-shot \\

\cmidrule(r){2-3} \cmidrule(r){4-5}

Traditional FSL &
$ 56.7  $ & $ 	75.93  $ & $ 	64.46  $ & $ 	82.21 $ \\

\midrule

Basic PS w/o PT &
$ 64.48  $ & $ 	71.41  $ & $ 	71.43  $ & $ 	77.79  $ \\
Masked PS w/o PT &
$ 67.37  $ & $ 	78.05  $ & $ 	83.81  $ & $90.55$ \\
\midrule

Basic PS  & $ 69.93 $ & $ 76.70 $ & $ 76.09  $ & $ 82.38$ \\
Masked PS &
$75.82$ & $85.93$ & $80.66$ & $87.83 $ \\

    \bottomrule
  \end{tabular}
\caption{EFSL results on \mini{} and \tiered{} with level 0 auxiliary data. Results are for the nearest centroid classifier, with various ablations and alternative setups. PT stands for pre-training in section two.  Accuracies are averaged over 800 episodes.}
\label{tab:transfer_part2}
\end{table}

%% file: tables/all_analysis.tex
\begin{table}[t]
  \centering
  \begin{tabular}{lcccc}
    \toprule
    
    Model & \multicolumn{2}{c}{\mini{}} & \multicolumn{2}{c}{\tiered{}} \\
    \midrule
    & 1-shot & 5-shot & 1-shot & 5-shot \\

\cmidrule(r){2-3} \cmidrule(r){4-5}

Traditional FSL &
$ 56.7  $ & $ 	75.93  $ & $ 	64.46  $ & $ 	82.21 $ \\
\midrule

MS K-Means~\citep{ren2018meta} &
$63.73$ & $83.97$ & $75.10$ & $87.28$ \\
\midrule

Basic Rnd PS &
$ 27.58 $ & $ 52.33 $ & $ 29.3 $ & $ 58.31 $ \\
Masked Rnd PS &
$ 48.87 $ & $ 	80.31 $ & $	68.21 $ & $	86.95 $ \\

\midrule

Masked PS &
$75.82$ & $85.93$ & $80.66$ & $87.83 $ \\

    \bottomrule
  \end{tabular}
\caption{EFSL results on \mini{} and \tiered{} with level 0 auxiliary data. Other than MS K-Means, the results are for the nearest centroid classifier. Accuracies are averaged over 800 episodes.}
\label{tab:ssfsl_random}
\end{table}

%% file: tables/imagenet_all_levels.tex
\begin{table*}[t]
  \centering
  \begin{tabular}{llccccccc}
    \toprule
    
    Embedding & Model & Level & \multicolumn{2}{c}{\mini{}} && \multicolumn{2}{c}{\tiered{}} \\
    \midrule
    &&& 1-shot & 5-shot && 1-shot & 5-shot \\
    
    \cmidrule(r){4-5} \cmidrule(r){7-8}

&NCC& \multirow{3}{*}{0} &   $ 69.93 \pm 0.76 $ & $ 76.70 \pm 0.64 $ && $ 76.09 \pm 0.81 $ & $ 82.38 \pm 0.67 $ \\
&RM &  &   $ 66.67 \pm 0.78 $ & $ 72.39 \pm 0.65 $ && $ 64.39 \pm 0.88 $ & $ 72.62 \pm 0.81 $ \\
&EA &  &   $ 72.42 \pm 0.72 $ & $ 80.58 \pm 0.58 $ && $ 75.52 \pm 0.80 $ & $ 81.94 \pm 0.69 $ \\
\cmidrule(r){2-8}

&NCC& \multirow{3}{*}{1} &   $ 68.82 \pm 0.74 $ & $ 74.98 \pm 0.64 $ && $ 72.17 \pm 0.93 $ & $ 80.37 \pm 0.78 $ \\
&RM &  &   $ 66.32 \pm 0.78 $ & $ 71.30 \pm 0.74 $ && $ 62.50 \pm 0.86 $ & $ 70.47 \pm 0.80 $ \\
&EA &  &   $ 70.58 \pm 0.72 $ & $ 76.98 \pm 0.67 $ && $ 71.74 \pm 0.87 $ & $ 80.51 \pm 0.76 $ \\
\cmidrule(r){2-8}

Basic PS &NCC& \multirow{3}{*}{2} &   $ 65.24 \pm 0.71 $ & $ 72.49 \pm 0.63 $ && $ 67.68 \pm 0.94 $ & $ 77.35 \pm  0.73 $ \\
&RM &  &   $ 62.30 \pm 0.76 $ & $ 69.01 \pm 0.67 $ && $ 51.80 \pm 0.81 $ & $ 60.94 \pm 0.88 $ \\
&EA &  &   $ 66.67 \pm 0.72 $ & $ 75.74 \pm 0.65 $ && $ 65.99 \pm 0.92 $ & $ 76.22 \pm 0.77 $ \\
\cmidrule(r){2-8}

&NCC& \multirow{3}{*}{3} &   $ 59.28 \pm 0.78 $ & $ 69.25 \pm 0.68 $ && $ 64.43 \pm 0.87 $ & $ 76.03 \pm 0.73 $ \\
&RM &  &   $ 48.66 \pm 0.73 $ & $ 59.43 \pm 0.70 $ && $ 46.80 \pm 0.86 $ & $ 61.44 \pm 0.84 $ \\
&EA &  &   $ 60.02 \pm 0.90 $ & $ 71.81 \pm 0.73 $ && $ 63.67 \pm 0.87 $ & $ 75.84 \pm 0.73 $ \\

\midrule

&NCC& \multirow{3}{*}{0} & $ 75.82 \pm 0.62 $ & $ 85.93 \pm 0.42 $ && $ 80.66 \pm 0.68 $ & $ 87.83 \pm 0.51 $ \\
&RM &  & $ 67.48 \pm 0.72 $ & $ 78.67 \pm 0.53 $ && $ 67.51 \pm 0.85 $ & $ 78.70 \pm 0.69 $ \\
&EA &  & $ 76.88 \pm 0.67 $ & $ 86.28 \pm 0.41 $ && $ 80.54 \pm 0.69 $ & $ 87.53 \pm 0.53 $ \\
\cmidrule(r){2-8}

&NCC& \multirow{3}{*}{1} & $ 74.24 \pm 0.68 $ & $ 82.87 \pm  0.53 $ && $ 79.09 \pm 0.78 $ & $ 87.74 \pm 0.53 $ \\
&RM &  & $ 68.81 \pm 0.76 $ & $ 76.55 \pm 0.66 $ && $ 67.14 \pm 0.79 $ & $ 79.61 \pm 0.68 $ \\
&EA &  & $ 75.43 \pm 0.69 $ & $ 83.98 \pm 0.51 $ && $ 80.54 \pm 0.69 $ & $ 87.53 \pm 0.53 $ \\
\cmidrule(r){2-8}

Masked PS &NCC& \multirow{3}{*}{2} & $ 71.60 \pm 0.64 $ & $ 82.49 \pm 0.47 $ && $ 73.82 \pm 0.74 $ & $ 86.75 \pm 0.51 $ \\
&RM &  & $ 65.76 \pm 0.69 $ & $ 77.28 \pm 0.57 $ && $ 62.02 \pm 0.88 $ & $ 77.12 \pm 0.75 $ \\
&EA &  & $ 73.79 \pm 0.66 $ & $ 83.84 \pm 0.49 $ && $ 73.29 \pm 0.76 $ & $ 86.36 \pm 0.53 $ \\
\cmidrule(r){2-8}

&NCC& \multirow{3}{*}{3} & $ 67.15 \pm 0.67 $ & $ 79.48 \pm 0.55 $ && $ 73.78 \pm 0.74 $ & $ 86.94 \pm 0.54 $ \\
&RM &  & $ 61.98 \pm 0.71 $ & $ 73.76 \pm 0.62 $ && $ 63.83 \pm 0.81 $ & $ 77.57 \pm 0.68 $ \\
&EA &  & $ 68.73 \pm 0.71 $ & $ 80.36 \pm 0.55 $ && $ 73.27 \pm 0.75 $ & $ 86.24 \pm 0.58 $ \\

    \bottomrule
  \end{tabular}
\caption{EFSL results on \mini{} and \tiered{} with level 0--3 auxiliary data. Results are for three classifier types: nearest-centroid classifier (NCC), relation module (RM), and embedding adaptation (EA). Each entry is mean accuracy over 800 episodes with 95\% confidence intervals.}
\label{tab:imagenet_all_levels}
\end{table*}

%% file: tables/cifar_all_levels.tex
\begin{table*}[t]
  \centering
  \begin{tabular}{llccccccc}
    \toprule
    
    Embedding & Model & Level & \multicolumn{2}{c}{\cifarfs{}} && \multicolumn{2}{c}{\fc{}} \\
    \midrule
    &&& 1-shot & 5-shot && 1-shot & 5-shot \\
    
    \cmidrule(r){4-5} \cmidrule(r){7-8}

&NCC& \multirow{3}{*}{0} &  $ 75.78 \pm 0.79 $ & $ 85.79 \pm 0.54 $ && $ 49.51 \pm 0.67 $ & $ 57.51 \pm 0.66 $ \\
&RM &  &   $ 73.32 \pm 0.84 $ & $ 81.68 \pm 0.64 $ && $ 43.22 \pm 0.65 $ & $ 49.18 \pm 0.66 $ \\
&EA &  &   $ 77.74 \pm 0.77 $ & $ 87.08 \pm 0.53 $ && $ 51.23 \pm 0.65 $ & $ 58.13 \pm 0.65 $ \\
\cmidrule(r){2-8}

&NCC& \multirow{3}{*}{1} &  $ 67.39 \pm 0.98 $ & $ 81.46 \pm 0.69 $ && $ 43.55 \pm 0.76 $ & $ 52.68 \pm 0.71 $ \\
&RM &  &   $ 63.96 \pm 0.95 $ & $ 77.41 \pm 0.71 $ && $ 38.49 \pm 0.75 $ & $ 45.68 \pm 0.69 $ \\
&EA &  &   $ 70.78 \pm 0.88 $ & $ 82.70 \pm  0.68 $ && $ 46.71 \pm 0.77 $ & $ 55.44 \pm 0.70 $ \\
\cmidrule(r){2-8}

Basic PS &NCC& \multirow{3}{*}{2} &  $ 56.66 \pm 1.06 $ & $ 75.55 \pm 0.69 $ && $ 36.04 \pm 0.75 $ & $ 48.85 \pm 0.72 $ \\
&RM &  &   $ 41.33 \pm 0.90 $ & $ 65.64 \pm 0.98 $ && $ 31.95 \pm 0.62 $ & $ 40.07 \pm 0.69 $ \\
&EA &  &   $ 56.16 \pm 0.95 $ & $ 75.75 \pm 0.66 $ && $ 37.64 \pm 0.78 $ & $ 50.40 \pm 0.70 $ \\
\cmidrule(r){2-8}

&NCC& \multirow{3}{*}{3} &  $ 47.56 \pm 1.00 $ & $ 73.71 \pm 0.73 $ && $ 30.00 \pm  0.63 $ & $ 43.12 \pm  0.64 $ \\
&RM &  &   $ 44.40 \pm 0.88 $ & $ 66.80 \pm 0.94 $ && $ 26.63 \pm 0.62 $ & $ 35.86 \pm 0.64 $ \\
&EA &  &   $ 49.32 \pm 0.94 $ & $ 74.48 \pm 0.72 $ && $ 30.58 \pm 0.64 $ & $ 43.47 \pm 0.62 $ \\
\midrule

&NCC& \multirow{3}{*}{0} &  $ 82.00 \pm 0.65 $ & $ 90.49 \pm 0.49 $ && $ 53.72 \pm 0.67 $ & $ 65.52 \pm  0.62 $ \\
&RM &  &   $ 74.17 \pm 0.80 $ & $ 86.21 \pm 0.57 $ && $ 42.04 \pm 0.65 $ & $ 52.02 \pm 0.63 $ \\
&EA &  &   $ 83.02 \pm 0.62 $ & $ 90.84 \pm 0.48 $ && $ 54.59 \pm 0.66 $ & $ 65.64 \pm 0.63 $ \\
\cmidrule(r){2-8}

&NCC& \multirow{3}{*}{1} &  $ 78.98 \pm 0.69 $ & $ 88.88 \pm 0.53 $ && $ 49.69 \pm 0.74 $ & $ 63.75 \pm 0.63 $ \\
&RM &  &   $ 74.55 \pm 0.79 $ & $ 82.96 \pm 0.66 $ && $ 40.47 \pm 0.68 $ & $ 51.82 \pm 0.65 $ \\
&EA &  &   $ 83.02 \pm 0.62 $ & $ 90.84 \pm 0.48 $ && $ 54.59 \pm 0.66 $ & $ 65.64 \pm 0.63 $ \\
\cmidrule(r){2-8}

Masked PS &NCC& \multirow{3}{*}{2} &  $ 69.90 \pm 0.80 $ & $ 84.69 \pm 0.54 $ && $ 49.12 \pm 0.70 $ & $ 64.12 \pm 0.67 $ \\
&RM &  &   $ 68.45 \pm 0.89 $ & $ 81.42 \pm 0.58 $ && $ 37.54 \pm 0.64 $ & $ 52.03 \pm 0.59 $ \\
&EA &  &   $ 70.07 \pm 0.79 $ & $ 84.59 \pm 0.56 $ && $ 49.32 \pm 0.70 $ & $ 63.94 \pm 0.68 $ \\
\cmidrule(r){2-8}

&NCC& \multirow{3}{*}{3} &  $ 68.53 \pm 0.72 $ & $ 84.68 \pm 0.55 $ && $ 41.72 \pm 0.60 $ & $ 57.72 \pm 0.59 $ \\
&RM &  &   $ 67.12 \pm 0.77 $ & $ 81.26 \pm 0.55 $ && $ 35.03 \pm 0.59 $ & $ 49.11 \pm 0.57 $ \\
&EA &  &   $ 69.26 \pm 0.70 $ & $ 84.65 \pm 0.54 $ && $ 41.53 \pm 0.58 $ & $ 57.26 \pm 0.61 $ \\

    \bottomrule
  \end{tabular}
\caption{EFSL results on \cifarfs{} and \fc{} with level 0--3 auxiliary data. Results are for three classifier types: nearest-centroid classifier (NCC), relation module (RM), and embedding adaptation (EA). Each entry is mean accuracy over 800 episodes with 95\% confidence intervals.}
\label{tab:cifar_all_levels}
\end{table*}